\pdfoutput=1
\documentclass[11pt]{article}

\usepackage[hyperref]{acl}
\usepackage{times}
\usepackage{latexsym}

\usepackage[T1]{fontenc}

\usepackage[utf8]{inputenc}
\usepackage[english]{babel}

\usepackage{microtype}
\usepackage{xcolor}
\usepackage{scalefnt}
\usepackage{amsmath}
\usepackage{graphicx}
\usepackage{multirow}
\usepackage{rotating}
\usepackage{tipa}
\usepackage{subfig}
\newcommand{\STAB}[1]{\begin{tabular}{@{}c@{}}#1\end{tabular}}
\usepackage{booktabs}
\usepackage{pgffor}

\title{``splink'' is happy and ``phrouth'' is scary:\\ Emotion Intensity Analysis for Nonsense Words}

\author{Valentino Sabbatino, Enrica Troiano, Antje Schweitzer, \and Roman Klinger\\
  Institut fur Maschinelle Sprachverarbeitung, University of Stuttgart, Germany\\
  \texttt{\{firstname.lastname\}@ims.uni-stuttgart.de}\\}

\begin{document}
\maketitle
\begin{abstract}
  People associate affective meanings to words -- ``death'' is scary
  and sad while ``party'' is connotated with surprise and joy. This
  raises the question if the association is purely a product of the
  learned affective imports inherent to semantic meanings, or is also
  an effect of other features of words, e.g., morphological and
  phonological patterns. We approach this question with an
  annotation-based analysis leveraging nonsense words.  Specifically,
  we conduct a best-worst scaling crowdsourcing study in which
  participants assign intensity scores for joy, sadness, anger,
  disgust, fear, and surprise to 272 nonsense words and, for
  comparison of the results to previous work, to 68 real words.  Based
  on this resource, we develop character-level and phonology-based
  intensity regressors. We evaluate them on both nonsense words
  and real words (making use of the NRC emotion intensity
  lexicon of 7493 words),
  across six emotion categories.  The analysis of our data reveals that some
  phonetic patterns show clear differences between emotion
  intensities. For instance, \textit{s} as a first phoneme contributes
  to joy, \textit{sh} to surprise, \textit{p} as last phoneme more to
  disgust than to anger and fear.  In the modelling experiments, a
  regressor trained on real words from the NRC emotion intensity
  lexicon shows a higher performance (\textit{r} = 0.17) than
  regressors that aim at learning the emotion connotation purely from
  nonsense words.  We conclude that humans do associate affective
  meaning to words based on surface patterns, but also based on
  similarities to existing words (``juy'' to ``joy'', or ``flike'' to
  ``like'').
\end{abstract}

\section{Introduction}
With words come meanings, as well as a variety of associations such
as emotional nuances. Emotions, feelings, and
attitudes, which can be summarized under the umbrella term of
``affect'', are in fact a core component for the meaning of large 
portions of a language vocabulary~\citep{Mohammad2018}. In English,
they encompass nouns, verbs, adjectives, and adverbs
 \citep{Mohammad2013}. 
For instance, \textit{dejected} and
\textit{wistful} can be said to directly express an emotion, but there 
are also terms that do
not describe a state of emotion and are still
associated to one (e.g., \textit{failure} and
\textit{death}\footnote{Examples from \citet{Mohammad2018}.}), 
given an interpretation of an associated
event.

Most computational studies of emotions in text deal with words in
context, for instance in news headlines 
\cite{Strapparava2007,Bostan2020} or in Tweets
\cite{schuff2017,Mohammad-2012,Koeper2017,prayas2017}.
Analyzing words in isolation, however, is equally important, as it can
help to create lexical resources for use in applications
\cite{Mohammad2013,Mohammad2018,Warriner2013}, 
to investigate how words are processed in general
\cite[Part 2]{Traxler2006}, and more specifically, to obtain a better
understanding of first language acquisition processes
\cite{bakhtiar2007nonword}.

When considering words in isolation, their meaning cannot be
disambiguated by the surrounding text. This raises the question: can
readers interpret an emotional load from unknown words, which are
judged out of their context? We address this question by analyzing
emotion associations of ``nonsense'' words -- or nonwords, or
pseudowords, i.e., terms which resemble real entries in the English
vocabulary, but are actually not part of it
\citep{keuleers2010wuggy,chuang2020processing}. Our aim is to
understand the degree to which nonsense words like \textit{fonk},
\textit{knunk}, or \textit{snusp} can be associated to particular
emotions. We model the problem as an emotion intensity analysis task
with a set of basic emotions, namely \textit{fear}, \textit{anger},
\textit{joy}, \textit{disgust}, \textit{surprise}, and
\textit{sadness}.

Other fields have provided evidence that some phonemes can be related
to the affective dimension of valence
\cite{myers2013inherent,adelman2018emotional}, but emotion analysis,
and in particular word-based research, has not yet ventured this
direction. Gaining insight on the emotional tone of non-existing
expressions could be relevant for current computational emotion
classification and intensity regression efforts, which have manifold
applications across social media mining or digital humanities. As an
example, when new product names are coined which do not have an
established semantics, designers and marketing experts might want to
be aware of the potential affective connections that these evoke, and
avoid those with a negative impact.

Therefore, our main contributions are: (1) the creation of an emotion
intensity lexicon of 272 nonsense words (with in addition 68 real
words, for comparison to previous work), (2) the analysis of the
phonemes present in them (if pronounced as English words) that aligns
with emotion intensity studies across the \citet{ekman1999basic} basic
emotions, and (3) experiments in which we develop intensity regressors
on a large resource of real words, as well as on our nonsense words. Both
regressors are evaluated on real and nonsense words.

\section{Related Work}
\subsection{Emotion Analysis}
Emotion analysis in text deals with the task of assigning (a set of)
emotions to words, sentences, or
documents~\citep{klinger2018analysis,schuff2017}, and is conducted
with various textual domains, including product reviews, tales, news,
and (micro)blogs~\citep{ea1, schuff2017}.  This task plays an
important role in applications like dialog systems (e.g., chatbots),
intelligent agents~\citep{klinger2018analysis} and for identifying
authors' opinions, affective intentions, attitudes, evaluations, and
inclinations~\citep{ea1}. Its scope extends beyond computer science 
and is of great interest for many fields,
like psychology, health care, and communication~\citep{ea2}.

Computational studies build on top of emotion theories in
psychology~\citep{ekman1999basic,plutchik2001nature,
  scherer2005emotions,russell1980circumplex}. While these theories by
and large agree that emotions encompass expressive,
behavioral, physiological, and phenomenological features, in emotion analysis 
they mainly serve as a reference system consisting of basic
emotions \citep{ekman1999basic,plutchik2001nature} or of a vector space
within which emotions can be represented
\citep{russell1980circumplex,scherer2005emotions}.

With respect to basic emotion approaches, dimensional ones explain
relations between emotions. The task of emotion intensity regression
can be thought of as a combination of these two. There, the goal is
not only to detect a categorical label, but also to recognize the
strength with which such emotion is expressed. This idea motivated a
set of shared tasks
\citep{MohammadB17wassa,mohammad-etal-2018-semeval}, some lexical
resources which assign emotion intensities to words
\citep{Mohammad2018} or to longer textual instances
\citep{mohammadb17starsem}, and automatic systems relying on deep
learning and said resources
\citep[i.a.]{prayas2017,Koeper2017,duppada2017seernet}.

\subsection{Nonsense Words and Emotional Sound Symbolism}
\label{sec:soundsymbolism}
Meaning in a language is 
conveyed in many different ways. At a phonetic level,
for example, languages systematically
use consonant voicing (/b/ vs. /p/, /d/ vs. /t/) to signal differences
in mass, vowel quality to signal size, vowel lengthening to signal
duration and intensity, reduplication to signal repetition, and in
some languages vowel height or frontality to mark
diminutives~\citep{davis2019horgous}.

Semantics has also been studied with respect to non-existing words
(i.e., terms without an established meaning).  By investigating their
lexical category, \citet{cassani2020semantics} explored the hypothesis
that there is ``(at least partially) a systematic relationship between
word forms and their meanings, such that children can infer'' the core
semantics of a word from its sound alone.  Also
\citet{chuang2020processing2} found that nonwords are semantically
loaded, and that their meanings co-determine lexical processing.
Their results indicate that ``non-word processing is influenced not
only by form similarity [..]  but also by nonword semantics''.

These ``nonsense meanings'' go beyond onomatopoeic connections:
\citet{cassani2020semantics} showed that high vowels tend to evoke
small forms, while low vowels tend to be associated with larger
forms. As a matter of facts, research has unvealed
many other links between visual and audio features
of stimuli,
besides the correspondences between verbal material and
the size of non-speech percepts. The loudness of sounds
and brightness of light
have been shown to be perceived similarly, at various degrees
of intensity \cite{bond1969cross}, and so are
pitch and visual brightness -- with higher pitched sounds being matched to bright stimuli
both by adults
\cite{marks1987cross} and children
 \cite{mondloch2004small}.
These findings are related to the so-called Bouba-Kiki
effect~\citep[p.\ 224]{Kohler1970} which describes a non-arbitrary
mapping between speech sounds and the visual shape of objects:
speakers in several languages pair nonsense words such as
\textit{maluma} or \textit{bouba} with round shapes, and
\textit{takete} or \textit{kiki} with spiky
ones~\citep{dOnofrio2014phonetic}.

Previous work exists also on the emotional connotation of word
sounds. \citet{majid2012current} provide an extensive overview of how
emotions saturate language at all levels, from prosody and the use of
interjections, to morphology and metaphoric expressions.  In
phonetics, the relationship between acoustic and affective phenomena
is based on the concept of sound
symbolism. \citet{adelman2018emotional} hypothesized that individual
phonemes are associated with negative and positive emotions and showed
that both phonemes at the beginning of a word and phonemes that are
pronounced fast convey negativity.  They demonstrated that emotional
sound symbolism is front-loaded, i.e., the first phoneme contributes
the most to decide the valence of a word.  Similarly,
\citet{myers2013inherent} showed that certain strings of English
phonemes have an inherent valence that can be predicted based on
dynamic changes in acoustic features.

In contrast to past research on emotional sound symbolism, ours
focuses on written material. In particular, we address nonsense words,
which are sequences of letters composing terms that do not exist in a
language~\citep{keuleers2010wuggy,chuang2020processing}, but conform
to its typical orthographic and phonological
patterns~\citep{keuleers2010wuggy}.  For this reason, they are of
particular interest in the psycholinguistics of language comprehension
\citep{bakhtiar2007nonword, keuleers2010wuggy, chuang2020processing,
  chuang2020processing2}.

\section{Data Acquisition and Annotation}
\label{DA}
We now describe the creation of our corpus of nonsense and real words,
with their respective emotion intensity scores for the six emotions of
\textit{joy}, \textit{sadness}, \textit{anger}, \textit{fear},
\textit{disgust}, and \textit{surprise}.\footnote{Our corpus is
  available base64 encoded in Appendix~\ref{sec:base}, and at
  \url{https://www.ims.uni-stuttgart.de/data/emotion}} We show an
excerpt of our data in Appendix~\ref{sec:corpusexcerpt}.

\subsection{Term Selection}
Our corpus consists of 272 nonsense words and 68 real words. The nonsense
words are taken from the ARC Nonword
Database\footnote{\url{http://www.cogsci.mq.edu.au/research/resources/nwdb/nwdb.html}}
\citep{rastle2002358}, which consists of 358,534 monosyllabic nonwords,
48,534 pseudohomophones, and 310,000 non-pseudohomophonic nonwords.
We randomly select nonsense words that have only orthographically
existing onsets and bodies and only monomorphemic syllables, such as
\textit{bleve}, \textit{foathe}, \textit{phlerm}, and \textit{snusp}.

In addition, for comparison to previous emotion intensity studies, we
sample a small number of words that are only linked to one emotion
from the NRC Emotion
Lexicon~\citep[EmoLex,][]{mohammadturney2010emotions}. This
resource contains a list of more than
$\approx$10k English words and their associations with eight emotions:
\textit{anger}, \textit{fear}, \textit{anticipation}, \textit{trust},
\textit{surprise}, \textit{sadness}, \textit{joy}, and
\textit{disgust}. 
Its creators outlined some best practices
to adopt in a crowdsourcing setup. They suggested to
collect judgments by asking workers if
a term is \textit{associated} to an emotion, as to obtain
more consistent judgments than could be collected by
asking whether the term \textit{evokes} an emotion.
We hence align with such strategy in the
design of our guidelines.

\begin{figure}
  \centering
  \includegraphics[width=\linewidth,height=6.5cm]{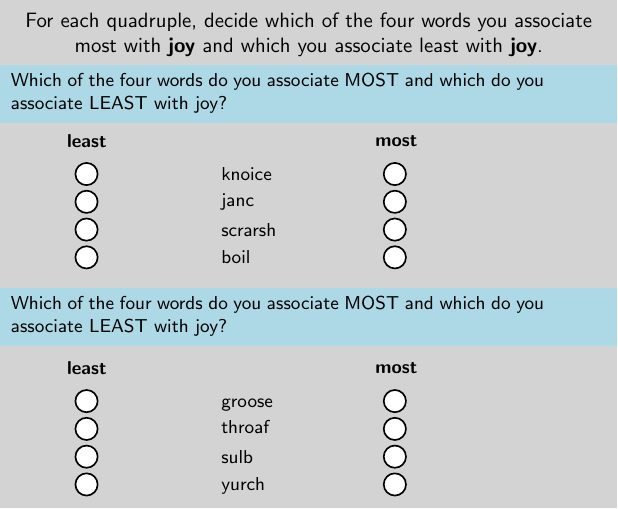}
  \caption{BWS Annotation Question example.}
  \label{fig:exampleannotation}
\end{figure}

\subsection{Annotation}
To obtain continuous intensity scores for each of the six emotions for each word,
we perform a best-worst scaling annotation~\citep[BWS,][]{louviere2015best,
  Mohammad2018} via crowdsourcing.

\paragraph{Study Setup.}
We follow the experimental setup described by \citet{bwsnaacl2016}.
For each experiment (i.e., an annotation task performed by three different annotators), 
we select N words out of the pool of 340 collected items. With these N words, we
randomly generate 2N distinct 4-tuples that comply with the
constraints of a word appearing in eight different tuples and no word
appearing in one tuple more than once.  We do this for all six
emotions. Therefore, each word occurs in 24 best-worst judgements
(8$\times$4$\times$3). Figure~\ref{fig:exampleannotation} exemplifies
the annotation task.

To aggregate the annotations to $\textrm{score}(w)$ for word $w$, we
take the normalized difference between the frequency with which the word was labeled
as best and as worst, i.e.,
$\textrm{score}(w)=\frac{\textrm{\#best}(w) -
  \textrm{\#worst}(w)}{\textrm{\#annotations}(w)}$~\citep{bwsnaacl2016}.
We linearly transform the score to $[0;1]$\footnote{We use an
  adaptation of the scripts from
  \url{http://saifmohammad.com/WebPages/BestWorst.html}}.

\paragraph{Attention Checks.}
To ensure annotation quality, we include attention checks. Each check
consists of an additional 4-tuple of only real, manually selected
words for the emotion in question. Two of the words are neutral with
respect to such emotion, and two are, respectively, strongly related
and opposite to it. For instance, we check attendance for \emph{joy}
with the words \textit{door}, \textit{elbow}, \textit{happiness}, and
\textit{depression}. Annotations by participants who fail any
attention check are discarded from our data.

\begin{table}
  \small\centering
  \begin{tabular}{lrrr}
    \toprule
    & Round 1 & Round 2 & Total \\
    \cmidrule(r){1-1}\cmidrule(rl){2-2}\cmidrule(rl){3-3}\cmidrule(l){4-4}
    \# Participants & 33 & 87 & 120 \\
    \mbox{}\hfill male & 11 & 19 & 30 \\
    \mbox{}\hfill female & 22 & 66 & 88 \\
    \mbox{}\hfill other &    & 2  & 2 \\
    \cmidrule(r){1-1}\cmidrule(rl){2-2}\cmidrule(rl){3-3}\cmidrule(l){4-4}
    Age & 31 & 32 & 31.5 \\
    \mbox{}\hfill min & 18 & 18 & 18 \\
    \mbox{}\hfill max & 61 & 65 & 65 \\
    \cmidrule(r){1-1}\cmidrule(rl){2-2}\cmidrule(rl){3-3}\cmidrule(l){4-4}
    \# Words & 55 & 290 & 340 \\
    \mbox{}\hfill non-words  & 44 & 232 & 272 \\
    \mbox{}\hfill real words & 11 & 58  & 68  \\
    \cmidrule(r){1-1}\cmidrule(rl){2-2}\cmidrule(rl){3-3}\cmidrule(l){4-4}
    Avg.\ duration & 15 min & 25 min & 20 min \\
    Overall cost & \textsterling 90.09  & \textsterling 395.85 & \textsterling 485.94 \\
    \bottomrule
  \end{tabular}
  \caption{Summary of the annotation study. The total
    number of words is 340 instead of 345, due to an overlap in 5
    selected words for Round 2.}
  \label{sum_rounds}
\end{table}

\subsubsection{Study Details}
Table~\ref{sum_rounds} summarizes the study details.  We hosted it on
the platform SoSci-Survey\footnote{\url{https://www.soscisurvey.de/}}
and recruited participants via
Prolific\footnote{\url{https://www.prolific.co/}}, rewarding them with
an hourly wage of \textsterling 7.80. We performed the annotations in
two iterations, the first of which was a small pretest to ensure the
feasibility of the task. In the second round, we increased the amount
of quadruples that one participant saw in one batch in each
experiment, i.e. from five words (four nonsensical ones) to 10 (eight
of which are nonsense).

Altogether, 120 participants worked on our 40 experiments, leading to
a total of 340 annotated words\footnote{A mistake in the word
  selection process led to an overlap of words, therefore we did not
  achieve 345 words but 340 words. We ignore the annotations of the
  affected tuples.}.  We prescreened participants to be native English
speakers and British citizens. Nevertheless, 19 participants indicated
in the study that they have a language proficiency below a native
speaker. All participants stated that they prefer British spelling
over other variants. 58 participants have a high school degree or
equivalent, 49 have a bachelor's degree, 11 have a master's degree and
2 have no formal qualification.

When asked for feedback regarding the study, participants remarked
that words with k's or v's sounded harsher and unfriendlier than
others, and expressed concern that assumptions about the pronunciation
of the judged terms might vary from person to person.  One participant
noticed that some nonsense words included familiar and existing words,
e.g., \textit{nice} in \textit{snice}, and this may have had an impact
on their choices.

\section{Corpus Analysis}
\label{CA}
We now discuss the reliability of the annotation
process and then analyze the resulting resource.

\begin{table}
  \centering\small
  \begin{tabular}{lrrrrrr}
    \toprule
    &\multicolumn{2}{c}{Nonsense}&\multicolumn{2}{c}{Real}&\multicolumn{2}{c}{NRC\,AIL}\\
    \cmidrule(lr){2-3}\cmidrule(lr){4-5}\cmidrule(l){6-7}
    Emotion & $\rho$ & $r$   & $\rho$& $r$   & $\rho$ & $r$ \\
    \cmidrule(r){1-1}\cmidrule(lr){2-3}\cmidrule(lr){4-5}\cmidrule(l){6-7}
    joy      & .68 & .72 & .87 & .87 & .93 & .92 \\
    sadness  & .62 & .68 & .87 & .88 & .90 & .91 \\
    anger    & .69 & .71 & .81 & .82 & .91 & .91 \\
    disgust  & .68 & .72 & .83 & .85 & --- & --- \\
    fear     & .65 & .70 & .82 & .85 & .91 & .91 \\
    surprise & .58 & .60 & .66 & .71 & --- & --- \\
    \hline
  \end{tabular}
  \caption{Split-half reliability for our nonsense word annotation
    in comparison to our real-word annotations and the
    scores obtained by \citet{Mohammad2018} (whose lexicon contains
    four out of our six emotions). $\rho$: Spearman correlation, $r$:
    Pearson correlation.}
  \label{shr_nonsense}
\end{table}

\subsection{Reliability and Distribution}
To assess the quality and reproducibility of our best-worst-scaling
annotations, we calculate split-half reliability\footnote{We use
  available implementations from \citet{bwsnaacl2016}:
  \url{http://saifmohammad.com/WebPages/BestWorst.html}.} (SHR) for
each emotion and summarize the results in Table~\ref{shr_nonsense}.
We observe that Spearman's $\rho$ correlation values for the nonsense
words are consistently below our real word annotations, with
differences between .08 and .25 points. Still, numbers 
indicate that annotations are strongly correlated.

Similar patterns hold for Pearson's $r$.
\textit{Sadness} shows the highest $r$ variation between the annotation of real 
and nonsense words ($r$=.88 vs .68); the emotion \textit{surprise} shows the smallest 
difference ($r$=.71 vs .60), 
but the absolute values of such correlations also lower than 
those obtained for other emotions.

\begin{figure}
  \centering
  \includegraphics[width=\columnwidth]{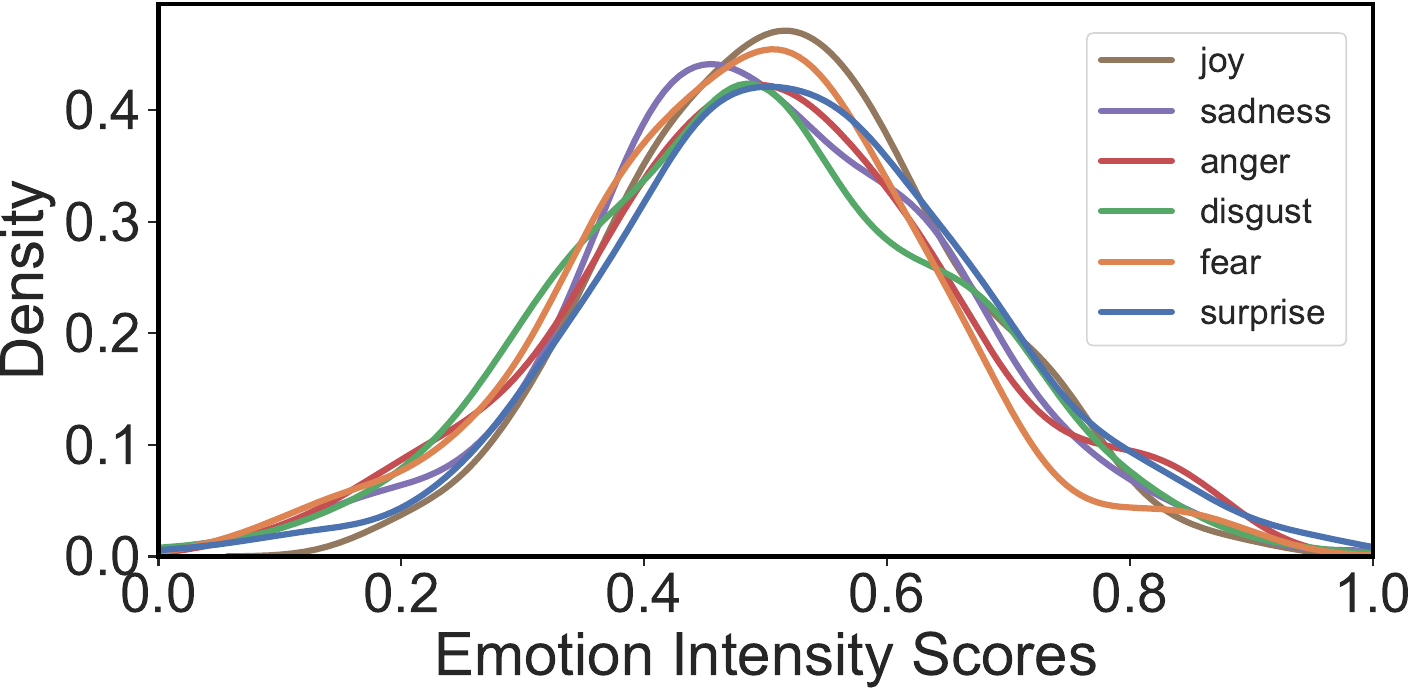}
  \caption{Density curves of nonsense word emotion intensities for our six emotions.}
  \label{fig:density}
\end{figure}

To compare these results to past research, we observe our
real word reliability scores to those found in work
describing the NRC lexicon (column NRC\,AIL in Table~\ref{shr_nonsense}).
Similar to such work, we also obtained highest results
for \textit{joy} than for emotions like \textit{anger} and
\textit{fear}. However, their results are generally higher, which
might be an effect of dataset size, and accordingly, a potentially
better familiarization of their annotators with the task.
Figure~\ref{fig:density} shows the distribution of the emotion
intensity values.  The plots for all emotions are similar and follow a
Gaussian distribution.

\begin{table*}
  \centering
  \small
  \setlength{\tabcolsep}{8pt}
  \renewcommand{\arraystretch}{1.1}
  \begin{tabular}{lrlrlrlrlrlr}
    \toprule
    \multicolumn{2}{c}{Joy} & \multicolumn{2}{c}{Sadness} & \multicolumn{2}{c}{Anger} & \multicolumn{2}{c}{Disgust} & \multicolumn{2}{c}{Fear} & \multicolumn{2}{c}{Surprise} \\
    \cmidrule(r){1-2}\cmidrule(rl){3-4}\cmidrule(rl){5-6}\cmidrule(rl){7-8}\cmidrule(rl){9-10}\cmidrule(l){11-12}
    Word & Int. & Word & Int. & Word & Int. & Word & Int. & Word & Int.& Word & Int. \\
    \cmidrule(r){1-2}\cmidrule(rl){3-4}\cmidrule(rl){5-6}\cmidrule(rl){7-8}\cmidrule(rl){9-10}\cmidrule(l){11-12}
    juy & .958 & vomp & .896 & terve & .938 & druss & .875 & phrouth & 1.0 & throoch & .896 \\ 
    flike & .938 & phlump & .875 & shait & .875 & pheague & .865 & ghoothe & .875 & shrizz & .875 \\ 
    splink & .938 & dis & .865 & phrouth & .854 & boarse & .854 & boarse & .854 & shrier & .833 \\ 
    glaim & .875 & losh & .854 & broin & .813 & snulge & .854 & wrorgue & .854 & spreil & .813 \\ 
    roice & .854 & drasque & .833 & psench & .813 & foathe & .833 & drasque & .833 & strem & .813 \\ 
    shrizz & .854 & weathe & .833 & slanc & .813 & gneave & .833 & dwalt & .833 & swunt & .792 \\ 
    spreece & .854 & dwaunt & .813 & straif & .813 & gream & .833 & keff & .813 & kease & .771 \\ 
    snusp & .833 & phlerm & .792 & thwealt & .792 & phlerm & .833 & bange & .792 & purf & .771 \\ 
    spirp & .833 & phreum & .792 & zorce & .792 & phlonch & .833 & frete & .792 & bange & .750 \\ 
    drean & .813 & sout & .792 & boarse & .771 & vomp & .833 & psoathe & .771 & droosh & .750 \\ 
    \bottomrule
  \end{tabular}
  \caption{Top ten nonsense words, ordered by decreasing emotion
    intensity.}
  \label{tab:topten}	
\end{table*}

In Table~\ref{tab:topten}, we report the top ten nonsense words with the
highest emotion intensity values for each emotion. These suggest
some hypotheses relative to how annotators decide on the emotion intensity.
Orthographical similarity to words with a clear emotional
connotation might have led to the emotion association to the nonsense
words. For instance,
\textit{juy} and \textit{flike} resemble the words \textit{joy} and
\textit{like}. Other nonwords might be interpreted by means of onomatopoeic
associations that arguably evoke events, like \textit{throoch} or \textit{shrizz} for \textit{surprise}
and \textit{snulge} or \textit{druss} in \textit{disgust}.

Some of these items exemplify the
importance of the first phonemes, in agreement with earlier work (see
Section~\ref{sec:soundsymbolism}). \textit{Surprise}-bearing nonwords, for instance,
tend to start with /s/ or /sh/, while
the second or third phoneme is often an /r/ sound\footnote{We use 
  ARPAbet for indicating phonemes.}.  Examples for this
pattern are \textit{shrizz}, \textit{shrier}, \textit{spreil}, and
\textit{strem}.

In addition, we observe that there is a relationship between the words
for the emotions \textit{sadness}, \textit{anger}, \textit{disgust},
and \textit{fear}.  For the emotion pairs
\textit{sadness}--\textit{disgust}, \textit{anger}--\textit{fear}, and
\textit{disgust}--\textit{fear} we have Pearson correlation values
ranging from 0.57 to 0.60.  For all the other different pairings of
emotions the Pearson correlation value is in $[0; 0.5]$.  Furthermore,
we can observe that for these four emotions we have negative Pearson
correlation values when comparing them with joy.  The Pearson
correlation values here lie between $-0.49$ and $-0.68$, where the
correlation is lowest for \textit{joy}--\textit{sadness} with a value
of $-0.68$.

\paragraph{Details on BWS Reliability Calculation.}
Our study has 2N (for N nonwords) BWS questions, that is, 4-tuples per
emotion. Since each nonword occurs on average in eight 4-tuples, and
three different annotators evaluate the same words, each word is
involved in $8 \times 3$ = 24 best-worst judgments.  In contrast to
the study design of \citet{bwsnaacl2016}, who ensure that the same
tuple is evaluated by multiple annotators, in our setup the nonword
are the unit being evaluated by the three annotators (but the tuples
may differ for each of them).  For us, one particular tuple might be
annotated by less than three annotators.

Therefore, we compute the SHR by randomly placing one or two
annotations per tuple in one bin and the remaining ones, if any
exists, for the tuple in another bin.  Then, two sets of intensity
values (and rankings) are computed from the annotations in each of the
two bins.  This process is repeated 100 times, and the correlations
between the two sets of rankings and intensity values are averaged per
emotion~\citep{MohammadB17wassa}.

\begin{sidewaysfigure*}[p]
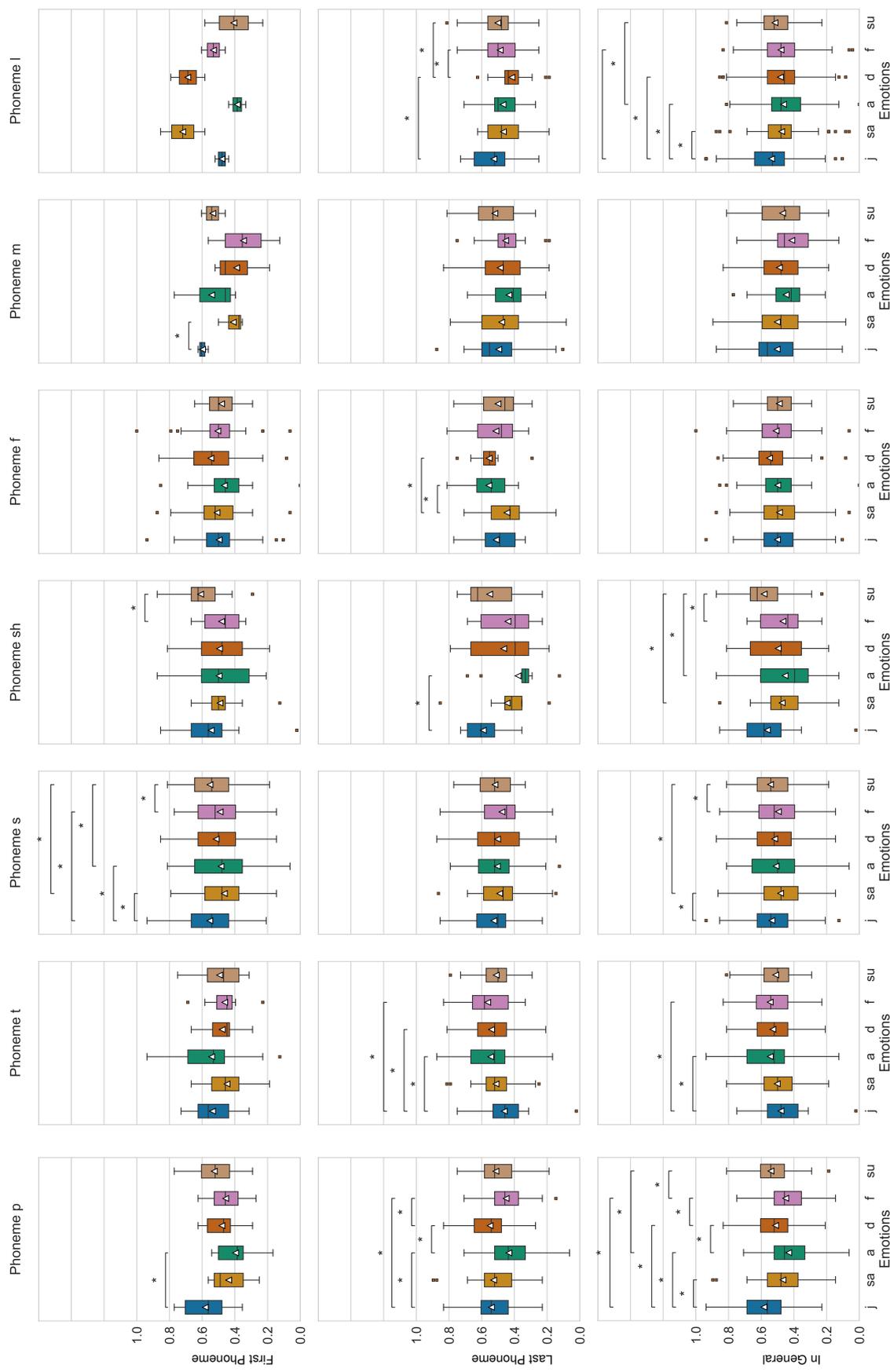

  \newcommand{\oneplotincTopRowFirst}[1]{\includegraphics[trim=0 0 35 0, clip, scale=0.28,page=#1]{fig/phonemes}}
  \newcommand{\oneplotincTopRowOther}[1]{\includegraphics[trim=40 0 35 0, clip, scale=0.28,page=#1]{fig/phonemes}}
  \newcommand{\oneplotincOtherRowFirst}[1]{\includegraphics[trim=0 0 35 65, clip, scale=0.28,page=#1]{fig/phonemes}}
  \newcommand{\oneplotincOtherRowOther}[1]{\includegraphics[trim=40 0 35 65, clip, scale=0.28,page=#1]{fig/phonemes}}
  \centering
  \rotatebox[y=50pt]{90}{\tiny\sf First Phoneme}\hspace{1pt}
  \oneplotincTopRowFirst{1}\hspace{4mm}
  \foreach \i in {2,...,7} {\oneplotincTopRowOther{\i}\hspace{4mm}}\\[-4mm]
  \rotatebox[y=50pt]{90}{\tiny\sf Last Phoneme}\hspace{1pt}
  \oneplotincOtherRowFirst{8}\hspace{4mm}
  \foreach \i in {9,...,14} {\oneplotincOtherRowOther{\i}\hspace{4mm}}\\[-4mm]
  \rotatebox[y=50pt]{90}{\tiny\sf In General}\hspace{1pt}
  \oneplotincOtherRowFirst{15}\hspace{4mm}
  \foreach \i in {16,...,21} {\oneplotincOtherRowOther{\i}\hspace{4mm}}\\[-4mm]
  \caption{Comparison of the emotion intensity distributions of phonemes /p/, /t/, /s/, /sh/, /f/, /m/, and /l/ 
    occurring as first or last phoneme in a word (rows one and two), or anywhere in a word (last row).
    The labels on the x-axis represent the emotions
    joy (j), sadness (sa), anger (a), disgust (d), fear (f), and
    surprise (su).  The asterisk (*) indicates p$\leq .05$, calculated
    with Welch's t-test between the intensity scores of the two
    emotions indicated by the bracket.}
  \label{fig:boxplots}
\end{sidewaysfigure*}

\subsection{Relation Between Phonemes and Emotion Intensities}
\label{Experiment_1}
Motivated by previous work on the emotional import of word sounds
\cite[e.g.,][]{adelman2018emotional}, we now analyse the relation
between specific phonemes and emotion intensities across our set of
emotions in our 272 annotated nonsense words.

\subsubsection{Experimental Setting}
For the phoneme analysis, we consider pronunciation, as it is provided
in the ARC Nonword Database. Pronounciation follows the DISC character
set of 42 symbols to represent 42
phonemes.\footnote{\url{https://www.cogsci.mq.edu.au/research/resources/nwdb/phonemes.html}}
We convert such representation to ARPAbet for consistency with real
word representations that are required for computational modelling
(see Section~\ref{Experiment_2}).

We focus on the three most frequent phonemes from each of the top 10
nonword lists in Table~\ref{tab:topten}. The selection results in the
eight phonemes /p/, /t/, /s/, /sh/, /f/, /m/, /l/, and
/r/.\footnote{Examples for these phonemes are /p/ as in \textit{pie},
  /t/ as in \textit{tie}, /s/ as in \textit{sigh}, /sh/ as in
  \textit{shy}, /f/ as in \textit{fight}, /m/ as in \textit{my}, /l/
  as in \textit{lie}, and /r/ as in \textit{rye}
  (\url{https://en.wikipedia.org/w/index.php?title=ARPABET&oldid=1062602312}).}
Next, we separate the words that have such phonemes in the first or
last position, or contain them in any position, and we compare the
distributions of their respective intensities for each emotion. We
calculate the p-values for the differences between the distributions
with Welch's t-test. We perform the t-test on sets of emotion
intensity scores that correspond to pairs of emotions, for the same
phoneme and the same position.

\subsubsection{Results}
Figure~\ref{fig:boxplots} illustrates the distributions of emotion
intensities for the chosen phonemes. The first row of plots
corresponds to the distribution for the subset of words in which the
phoneme appears in the first position of the nonword, the second row
to the appearance as a last phoneme, and the third row relates to
nonwords containing that phoneme at any possible position. Differences
between emotions that have a p-value below 0.05 are denoted with a
$*$. We limit our discussion to these cases.

\paragraph{1st Phoneme.} For the phonemes /p/, /s/, /sh/, and /m/,
certain emotion pairs show a p-value below 5\%. For /p/ and /s/,
\textit{joy} has the highest median intensity (as in \textit{splink},
\textit{spreece}, \textit{snusp}), and \textit{anger} the
lowest. Examples for low \textit{joy} intensities which still have an
/s/ at the beginning are \textit{slanc} or \textit{scunch} -- but
other parts of the nonword also seem to play an important role here.
\textit{Surprise} has a stronger intensity than all other emotions for
items with /sh/ in first position, particularly in comparison to
\textit{fear} (p$<$.05 only for \textit{joy/fear}). Examples for
strongly \textit{surprise}-loaded words are \textit{shrizz},
\textit{shrier}, and \textit{shoach}. Counterexamples are
\textit{shogue} and \textit{shuilt}.

Another noteworthy pattern is observable with the phoneme /m/, for
which \textit{joy} is substantially higher than \textit{sadness}. It
should be noted, however, that there are only three instances in our
dataset starting with /m/ (i.e., \textit{maut, marve, mauge}).

An interesting case is the occurrence of /t/ and its relation to
\textit{anger} intensities. These values cover a wide interval:
examples for high \textit{anger} degrees are \textit{terve},
\textit{trasque}, and \textit{tource}, low intensity ones are
\textit{tish} and \textit{twauve}. We hypothesize that the combination
of /t/ with /r/ might be relevant.

\paragraph{Last Phoneme.} Interestingly, and in contradiction to our
expectations based on previous work, the occurrences of last phonemes
of nonwords are related to a set of differences in emotion
intensities. For /p/, \textit{disgust} nonwords have the highest
intensity, being clearly different from \textit{anger} as well as
\textit{fear}, which are associated with comparably low values. /sh/,
which showed interesting patterns in the first phoneme relative to
\textit{surprise}, contributes most to \textit{joy} when found in the
last position (as in \textit{tish}), in contrast to instances that
evoke negative emotions like \textit{anger}.

\paragraph{General.} The analysis of phonemes independent of their
positions leads more often to comparably low p-values due to larger
numbers of words in each set. The patterns, however, by and large
resemble the observations for the first and the last phonemes.

\section{Modeling}
\label{Experiment_2}
Our analysis has revealed that particular phonemes are indeed related
to high intensities for some emotions. In the following section, we
aim at understanding if these findings are exploited by computational
models that perform emotion intensity regression (i.e., if these
models perform better when they observe specific character sequences
or phoneme sequences), and if a model that is trained on real words
can generalize the learned emotion associations to nonsense words (or
the other way around).

\subsection{Experimental Setting}
As for our architecture, we build on top of the model proposed by
\citet{Koeper2017} for Tweets. This model is a combination of a
convolutional neural network with a bidirectional long short-term
memory model. We opt against using a pre-trained transfomer approach
like BERT \cite{BERT}, to have full control over input sequences -- we
use character or phoneme sequences as input. These are represented as
300 dimensional embeddings, with the maximal sequence length being 16,
which corresponds to the longest input sequence in our corpus
(including real words from NRC-EIL, see below). We apply a dropout
rate of 0.25, convolutions with window size of 3, followed by a max
pooling layer of size 2 and a BiLSTM.

\paragraph{Train/Test Split.} We divide the 272 data points into a
train set of 204 nonsense words and a test set of 68 nonsense
words. We further use the NRC-EIL lexicon \cite{Mohammad2018} with
1268 words for \textit{joy}, 1298 for \textit{sadness}, 1483 for
\textit{anger}, 1094 for \textit{disgust}, 1765 for \textit{fear}, and
585 for \textit{surprise}. We also split this corpus into train/test
set, with 75\,\% of the data for training.

\paragraph{Phoneme Representation.} We represent both nonsense words
and real words as phoneme sequences following the ARPAbet
representation.  For the words from the NRC-EIL, we obtain the ARPAbet
pronunciation from the Carnegie Mellon University (CMU) Pronouncing
Dictionary (CMUdict). For words that are not included in CMUdict, we
use the LOGIOS Lexicon Tool, which adds normalization heuristics on
top of CMUdict.\footnote{CMUdict:
  \url{http://www.speech.cs.cmu.edu/cgi-bin/cmudict}, LOGIOS:
  \url{http://www.speech.cs.cmu.edu/tools/lextool.html}. Both URLs are
  not available as of April 2022. The websites can be accessed via the
  Wayback Machine at 
  \url{https://web.archive.org/web/20211109084743/http://www.speech.cs.cmu.edu/tools/lextool.html}
  and
  \url{https://web.archive.org/web/20210815020323/http://www.speech.cs.cmu.edu/tools/lextool.html}.}

\paragraph{Input Embeddings.}
We compare two input representations, character embeddings and phoneme
embeddings. For the character representations, we use pretrained
FastText embeddings, which provide character-level information. These
embeddings are trained on 400 million Tweets \cite{godin2019}. We
train the phoneme embeddings on the established corpus of 7392
sentences by \citet{phn2vec} which is based on the DARPA TIMIT
Acoustic-Phonetic Continuous Speech Corpus~\citep{timit}.

\begin{figure*}
  \centering
  \includegraphics[width=1.3\columnwidth]{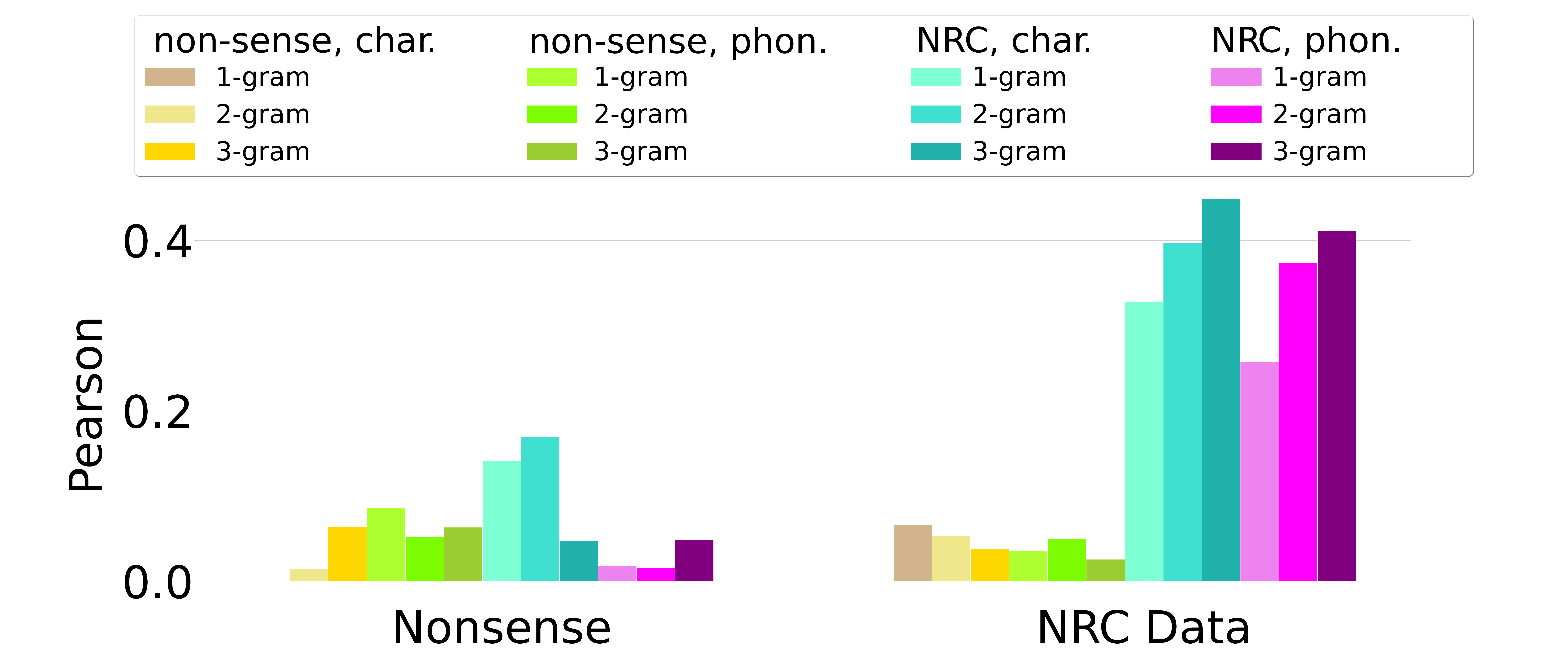}
  \caption{Barplot for Pearson correlation (averaged over all
    emotions). Each bar corresponds to one model configuration, either
    trained on nonsense words or on real words (NRC), with character
    embedding input or phoneme embedding input.}\label{avg_metrics}
\end{figure*}

\paragraph{Model Variants.}
We compare models that differ in the following parameters: (1) input
representation (characters/phonemes), (2) n-grams length over
characters/phonemes (1/2/3 grams), (3) input training data (real words
from NRC-EIL, our nonsense words).  The reason for considering
different n-grams is that, in addition to the standard use of
unigrams, we also want to investigate 2- and 3-grams under the
assumption that the inter-word relationship can be better captured
with n-grams.  The FastText embeddings provide the capability to work
with n-grams out-of-the-box.  We do not fine-tune the pre-trained
embeddings for the respective prediction task.

For each of the 12 models, we train a separate regressor per emotion,
as an alternative to multi-task models. This choice prevents the
output emotion labels from interacting in the intensity predictions.
Furthermore, preliminary experiments helped us establish that joint
multi-task models are inferior to single regressors for our task.
\subsection{Results}
Figure~\ref{avg_metrics} summarizes the results of our 12 emotion
intensity prediction models and presents the performance using Pearson
correlation ($r$).  Numbers are average values over the results
per emotion.

We first consider the models when tested on nonsense words (the left
12 bars in the figure). The phoneme-based models trained on nonsense
words show slightly higher performance than the character-based
models, but all these models are clearly outperformed by
character-based models trained on real words. Therefore, we conclude
that a model trained on real words does enable emotion intensity
prediction on nonsense words, though to a limited degree
($r$=0.17). This is in accordance with the fact that human annotators
declared to relate some of their judgments to existing English terms.

On the other side, testing on real words reveals a low performance of
the models that were trained on nonsense words: the meaning of real
words seems to dominate over phonetic patterns to take emotion
decisions, which is a type of information that cannot be relied upon
when training on nonwords. We should acknowledge, however, that this
setup provided the models with an exceptionally limited amount of
data, thus making it difficult to conclude that phonetic patterns do
not play any role in automatic emotion inferences.

\section{Conclusion \& Future Work}
\label{Conclusion}
We addressed the question of whether humans associate emotion
intensities with nonsense words and tested if machine learning-based
regressors pick up phonetic patterns to make emotion intensity
predictions.  Our annotation study revealed that humans do indeed make
such associations. Especially the first phoneme of a word influences
the resulting emotion intensity judgement: /p/ and /s/ seem to
increase the perception of \textit{joy}, /sh/ of \textit{surprise},
and /m/ is more likely related to \textit{sadness}.  Contrary to our
assumptions, phonemes placed at the last position of a nonword also
play an important role. The phoneme /p/, for instance, points towards
an increased degree of \textit{disgust}.

We found that our emotion intensity regressors do predict emotion
intensity based on word form and pronunciation, although only to a
limited degree for nonsense words. Training on nonsense items and
testing on real vocabulary entries results in a low performance, thus
indicating that the meaning of known words overrules patterns that can
be deduced from nonsense ones. When learned the other way around, our
computational models make use of patterns found in real words that, to
some degree, allow the emotion intensity prediction on nonsense
counterparts.

One limitation of this first study of written nonsense words and their
emotion association is the comparably limited size of the corpus we
compiled. Future work could perform the annotation study with more
items and across more diverse sets of annotators.  Furthermore, our
analysis focused on single phonemes that we selected based on their
frequency in the data. This way of selecting the phonemes under
investigation neglects the dependence between their frequencies and
their positions. It also disregards potential interactions between
different phonemes, as well as the role of less frequent phonemes in
emotion intensity decisions. Future work should take into account
these types of considerations.

\section*{Acknowledgements}
We thank Sebastian Pad\'o for helpful discussions. This research is
funded by the German Research Council (DFG), project ``Computational
Event Analysis based on Appraisal Theories for Emotion Analysis''
(CEAT, project number KL 2869/1-2).
 
\bibliography{lit}

\begin{thebibliography}{43}
\expandafter\ifx\csname natexlab\endcsname\relax\def\natexlab#1{#1}\fi

\bibitem[{Adelman et~al.(2018)Adelman, Estes, and Cossu}]{adelman2018emotional}
James~S. Adelman, Zachary Estes, and Martina Cossu. 2018.
\newblock \href
  {https://www.sciencedirect.com/science/article/pii/S0010027718300374}
  {{Emotional sound symbolism: Languages rapidly signal valence via phonemes}}.
\newblock \emph{Cognition}, 175:122--130.

\bibitem[{Aman and Szpakowicz(2007)}]{ea1}
Saima Aman and Stan Szpakowicz. 2007.
\newblock \href
  {https://link.springer.com/chapter/10.1007/978-3-540-74628-7_27} {Identifying
  expressions of emotion in text}.
\newblock In \emph{Text, Speech and Dialogue}, pages 196--205, Berlin,
  Heidelberg. Springer Berlin Heidelberg.

\bibitem[{Bakhtiar et~al.(2007)Bakhtiar, Abad~Ali, and
  Sadegh}]{bakhtiar2007nonword}
Mehdi Bakhtiar, Dehqan Abad~Ali, and Seif Sadegh. 2007.
\newblock \href {https://pubmed.ncbi.nlm.nih.gov/17679736/} {Nonword repetition
  ability of children who do and do not stutter and covert repair hypothesis}.
\newblock \emph{Indian Journal of Medical Sciences}, 61(8):462--470.

\bibitem[{Bond and Stevens(1969)}]{bond1969cross}
Barbara Bond and Stanley~S Stevens. 1969.
\newblock \href {https://doi.org/10.3758/BF03212787} {Cross-modality matching
  of brightness to loudness by 5-year-olds}.
\newblock \emph{Perception \& Psychophysics}, 6(6):337--339.

\bibitem[{Bostan et~al.(2020)Bostan, Kim, and Klinger}]{Bostan2020}
Laura Ana~Maria Bostan, Evgeny Kim, and Roman Klinger. 2020.
\newblock \href {https://aclanthology.org/2020.lrec-1.194}
  {{G}ood{N}ews{E}veryone: A corpus of news headlines annotated with emotions,
  semantic roles, and reader perception}.
\newblock In \emph{Proceedings of the 12th Language Resources and Evaluation
  Conference}, pages 1554--1566, Marseille, France. European Language Resources
  Association.

\bibitem[{Bostan and Klinger(2018)}]{klinger2018analysis}
Laura-Ana-Maria Bostan and Roman Klinger. 2018.
\newblock \href {https://aclanthology.org/C18-1179} {An analysis of annotated
  corpora for emotion classification in text}.
\newblock In \emph{Proceedings of the 27th International Conference on
  Computational Linguistics}, pages 2104--2119, Santa Fe, New Mexico, USA.
  Association for Computational Linguistics.

\bibitem[{Cassani et~al.(2020)Cassani, Chuang, and
  Baayen}]{cassani2020semantics}
Giovanni Cassani, Yu-Ying Chuang, and R~Harald Baayen. 2020.
\newblock \href {https://doi.org/10.1037/xlm0000747} {On the {S}emantics of
  {N}onwords and {Their} {Lexical} {C}ategory}.
\newblock \emph{Journal of Experimental Psychology: Learning, Memory, and
  Cognition}, 46(4):621--637.

\bibitem[{Chaffar and Inkpen(2011)}]{ea2}
Soumaya Chaffar and Diana Inkpen. 2011.
\newblock \href {https://link.springer.com/chapter/10.1007/978-3-642-21043-3_8}
  {Using a heterogeneous dataset for emotion analysis in text}.
\newblock In \emph{Advances in Artificial Intelligence}, pages 62--67, Berlin,
  Heidelberg. Springer Berlin Heidelberg.

\bibitem[{Chuang et~al.(2019)Chuang, Vollmer, Shafaei-Bajestan, Gahl, Hendrix,
  and Baayen}]{chuang2020processing2}
Yu-Ying Chuang, Marie-lenka Vollmer, Elnaz Shafaei-Bajestan, Susanne Gahl,
  Peter Hendrix, and R~Harald Baayen. 2019.
\newblock \href {https://assta.org/proceedings/ICPhS2019/papers/ICPhS_1282.pdf}
  {On the processing of nonwords in word naming and auditory lexical decision}.
\newblock In \emph{Proceedings of the 19th International Congress of Phonetic
  Sciences, Melbourne, Australia 2019}, pages 1233--1237. Australasian Speech
  Science and Technology Association Inc.

\bibitem[{Chuang et~al.(2021)Chuang, Vollmer, Shafaei-Bajestan, Gahl, Hendrix,
  and Baayen}]{chuang2020processing}
Yu-Ying Chuang, Marie~Lenka Vollmer, Elnaz Shafaei-Bajestan, Susanne Gahl,
  Peter Hendrix, and R~Harald Baayen. 2021.
\newblock \href {https://doi.org/10.3758/s13428-020-01356-w} {The processing of
  pseudoword form and meaning in production and comprehension: {A}
  computational modeling approach using linear discriminative learning}.
\newblock \emph{Behavior research methods}, 53(3):945--976.

\bibitem[{Davis et~al.(2019)Davis, Morrow, and Lupyan}]{davis2019horgous}
Charles~P. Davis, Hannah~M. Morrow, and Gary Lupyan. 2019.
\newblock \href {https://onlinelibrary.wiley.com/doi/abs/10.1111/cogs.12791}
  {What {Does} a {Horgous Look Like}? {N}onsense {Words Elicit Meaningful
  Drawings}}.
\newblock \emph{Cognitive Science}, 43(10):e12791.

\bibitem[{Devlin et~al.(2019)Devlin, Chang, Lee, and Toutanova}]{BERT}
Jacob Devlin, Ming-Wei Chang, Kenton Lee, and Kristina Toutanova. 2019.
\newblock \href {https://doi.org/10.18653/v1/N19-1423} {{BERT}: {Pre-training
  of Deep Bidirectional Transformers for Language Understanding}}.
\newblock In \emph{Proceedings of the 2019 Conference of the North {A}merican
  Chapter of the Association for Computational Linguistics: Human Language
  Technologies, Volume 1 (Long and Short Papers)}, pages 4171--4186,
  Minneapolis, Minnesota. Association for Computational Linguistics.

\bibitem[{Duppada and Hiray(2017)}]{duppada2017seernet}
Venkatesh Duppada and Sushant Hiray. 2017.
\newblock \href {https://doi.org/10.18653/v1/W17-5228} {Seernet at
  {E}mo{I}nt-2017: Tweet emotion intensity estimator}.
\newblock In \emph{Proceedings of the 8th Workshop on Computational Approaches
  to Subjectivity, Sentiment and Social Media Analysis}, pages 205--211,
  Copenhagen, Denmark. Association for Computational Linguistics.

\bibitem[{D’Onofrio(2014)}]{dOnofrio2014phonetic}
Annette D’Onofrio. 2014.
\newblock \href {https://doi.org/10.1177/0023830913507694} {{Phonetic Detail
  and Dimensionality in Sound-shape Correspondences: Refining the Bouba-Kiki
  Paradigm}}.
\newblock \emph{Language and Speech}, 57(3):367--393.

\bibitem[{Ekman(1999)}]{ekman1999basic}
Paul Ekman. 1999.
\newblock \href
  {https://onlinelibrary.wiley.com/doi/abs/10.1002/0470013494.ch3}
  {\emph{{Basic Emotions}}}, chapter~3. John Wiley \& Sons, Ltd.

\bibitem[{Garofolo et~al.(1993)Garofolo, Lamel, Fisher, Fiscus, Pallett,
  Dahlgren, and Zue}]{timit}
John~S. Garofolo, Lori~F. Lamel, William~M. Fisher, Jonathan~G. Fiscus,
  David~S. Pallett, Nancy~L. Dahlgren, and Victor Zue. 1993.
\newblock \href {https://doi.org/10.35111/17gk-bn40} {{TIMIT} acoustic-phonetic
  continuous speech corpus}.
\newblock Linguistic Data Consortium.

\bibitem[{Godin(2019)}]{godin2019}
Fr\'{e}deric Godin. 2019.
\newblock \href {https://biblio.ugent.be/publication/8622030} {\emph{Improving
  and Interpreting Neural Networks for Word-Level Prediction Tasks in Natural
  Language Processing}}.
\newblock Ph.D. thesis, Ghent University, Belgium.

\bibitem[{Goel et~al.(2017)Goel, Kulshreshtha, Jain, and Shukla}]{prayas2017}
Pranav Goel, Devang Kulshreshtha, Prayas Jain, and Kaushal~Kumar Shukla. 2017.
\newblock \href {https://aclanthology.org/W17-5207} {Prayas at {E}mo{I}nt 2017:
  {An Ensemble of Deep Neural Architectures for Emotion Intensity Prediction in
  Tweets}}.
\newblock In \emph{Proceedings of the 8th Workshop on Computational Approaches
  to Subjectivity, Sentiment and Social Media Analysis}, pages 58--65,
  Copenhagen, Denmark. Association for Computational Linguistics.

\bibitem[{Keuleers and Brysbaert(2010)}]{keuleers2010wuggy}
Emmanuel Keuleers and Marc Brysbaert. 2010.
\newblock \href {https://doi.org/10.3758/BRM.42.3.627} {{Wuggy: A multilingual
  pseudoword generator}}.
\newblock \emph{Behavior Research Methods}, 42(3):627--633.

\bibitem[{Kiritchenko and Mohammad(2016)}]{bwsnaacl2016}
Svetlana Kiritchenko and Saif~M. Mohammad. 2016.
\newblock \href {https://doi.org/10.18653/v1/N16-1095} {Capturing reliable
  fine-grained sentiment associations by crowdsourcing and best{--}worst
  scaling}.
\newblock In \emph{Proceedings of the 2016 Conference of the North {A}merican
  Chapter of the Association for Computational Linguistics: Human Language
  Technologies}, pages 811--817, San Diego, California. Association for
  Computational Linguistics.

\bibitem[{K\"ohler(1970)}]{Kohler1970}
Wolfgang K\"ohler. 1970.
\newblock \emph{Gestalt Psychology}.
\newblock Liveright, New York.

\bibitem[{K\"oper et~al.(2017)K\"oper, Kim, and Klinger}]{Koeper2017}
Maximilian K\"oper, Evgeny Kim, and Roman Klinger. 2017.
\newblock \href {https://www.aclanthology.org/W17-5206/} {{IMS} at
  {EmoInt-2017}: {Emotion Intensity Prediction with Affective Norms,
  Automatically Extended Resources and Deep Learning}}.
\newblock In \emph{Proceedings of the 8th Workshop on Computational Approaches
  to Subjectivity, Sentiment and Social Media Analysis}, Copenhagen, Denmark.
  Workshop at Conference on Empirical Methods in Natural Language Processing,
  Association for Computational Linguistics.

\bibitem[{Louviere et~al.(2015)Louviere, Flynn, and Marley}]{louviere2015best}
Jordan~J. Louviere, Terry~N. Flynn, and A.~A.~J. Marley. 2015.
\newblock \href {https://doi.org/10.1017/CBO9781107337855} {\emph{Best-Worst
  Scaling: Theory, Methods and Applications}}.
\newblock Cambridge University Press.

\bibitem[{Majid(2012)}]{majid2012current}
Asifa Majid. 2012.
\newblock \href {https://doi.org/10.1177/1754073912445827} {Current emotion
  research in the language sciences}.
\newblock \emph{Emotion Review}, 4(4):432--443.

\bibitem[{Marks(1987)}]{marks1987cross}
Lawrence~E. Marks. 1987.
\newblock \href {https://doi.org/10.1037//0096-1523.13.3.384} {On cross-modal
  similarity: Auditory--visual interactions in speeded discrimination.}
\newblock \emph{Journal of Experimental Psychology: Human Perception and
  Performance}, 13(3):384.

\bibitem[{Mohammad(2012)}]{Mohammad-2012}
Saif Mohammad. 2012.
\newblock \href {https://aclanthology.org/S12-1033} {{\#}emotional tweets}.
\newblock In \emph{*{SEM} 2012: The First Joint Conference on Lexical and
  Computational Semantics {--} Volume 1: Proceedings of the main conference and
  the shared task, and Volume 2: Proceedings of the Sixth International
  Workshop on Semantic Evaluation ({S}em{E}val 2012)}, pages 246--255,
  Montr{\'e}al, Canada. Association for Computational Linguistics.

\bibitem[{Mohammad(2018)}]{Mohammad2018}
Saif Mohammad. 2018.
\newblock \href {https://aclanthology.org/L18-1027} {Word affect intensities}.
\newblock In \emph{Proceedings of the Eleventh International Conference on
  Language Resources and Evaluation ({LREC} 2018)}, Miyazaki, Japan. European
  Language Resources Association (ELRA).

\bibitem[{Mohammad and Bravo-Marquez(2017{\natexlab{a}})}]{mohammadb17starsem}
Saif Mohammad and Felipe Bravo-Marquez. 2017{\natexlab{a}}.
\newblock \href {https://doi.org/10.18653/v1/S17-1007} {Emotion intensities in
  tweets}.
\newblock In \emph{Proceedings of the 6th Joint Conference on Lexical and
  Computational Semantics (*{SEM} 2017)}, pages 65--77, Vancouver, Canada.
  Association for Computational Linguistics.

\bibitem[{Mohammad and Bravo-Marquez(2017{\natexlab{b}})}]{MohammadB17wassa}
Saif Mohammad and Felipe Bravo-Marquez. 2017{\natexlab{b}}.
\newblock \href {https://doi.org/10.18653/v1/W17-5205} {{WASSA}-2017 shared
  task on emotion intensity}.
\newblock In \emph{Proceedings of the 8th Workshop on Computational Approaches
  to Subjectivity, Sentiment and Social Media Analysis}, pages 34--49,
  Copenhagen, Denmark. Association for Computational Linguistics.

\bibitem[{Mohammad et~al.(2018)Mohammad, Bravo-Marquez, Salameh, and
  Kiritchenko}]{mohammad-etal-2018-semeval}
Saif Mohammad, Felipe Bravo-Marquez, Mohammad Salameh, and Svetlana
  Kiritchenko. 2018.
\newblock \href {https://doi.org/10.18653/v1/S18-1001} {{S}em{E}val-2018 task
  1: Affect in tweets}.
\newblock In \emph{Proceedings of The 12th International Workshop on Semantic
  Evaluation}, pages 1--17, New Orleans, Louisiana. Association for
  Computational Linguistics.

\bibitem[{Mohammad and Turney(2010)}]{mohammadturney2010emotions}
Saif Mohammad and Peter Turney. 2010.
\newblock \href {https://aclanthology.org/W10-0204} {{Emotions Evoked by Common
  Words and Phrases: Using {M}echanical {T}urk to Create an Emotion Lexicon}}.
\newblock In \emph{Proceedings of the {NAACL} {HLT} 2010 Workshop on
  Computational Approaches to Analysis and Generation of Emotion in Text},
  pages 26--34, Los Angeles, CA. Association for Computational Linguistics.

\bibitem[{Mohammad and Turney(2013)}]{Mohammad2013}
Saif~M. Mohammad and Peter~D. Turney. 2013.
\newblock \href {https://doi.org/10.1111/j.1467-8640.2012.00460.x}
  {{Crowdsourcing a Word-Emotion Association Lexicon}}.
\newblock \emph{Computational Intelligence}, 29(3):436--465.

\bibitem[{Mondloch and Maurer(2004)}]{mondloch2004small}
Catherine~J. Mondloch and Daphne Maurer. 2004.
\newblock \href {https://doi.org/10.3758/CABN.4.2.133} {Do small white balls
  squeak? {Pitch-object} correspondences in young children}.
\newblock \emph{Cognitive, Affective, \& Behavioral Neuroscience},
  4(2):133--136.

\bibitem[{Myers-Schulz et~al.(2013)Myers-Schulz, Pujara, Wolf, and
  Koenigs}]{myers2013inherent}
Blake Myers-Schulz, Maia Pujara, Richard~C Wolf, and Michael Koenigs. 2013.
\newblock \href {https://doi.org/10.1080/02699931.2012.754739} {Inherent
  emotional quality of human speech sounds}.
\newblock \emph{Cognition and Emotion}, 27(6):1105--1113.

\bibitem[{Plutchik(2001)}]{plutchik2001nature}
Robert Plutchik. 2001.
\newblock \href {http://www.jstor.org/stable/27857503} {{The nature of
  emotions: Human emotions have deep evolutionary roots, a fact that may
  explain their complexity and provide tools for clinical practice}}.
\newblock \emph{American Scientist}, 89(4):344--350.

\bibitem[{Rastle et~al.(2002)Rastle, Harrington, and Coltheart}]{rastle2002358}
Kathleen Rastle, Jonathan Harrington, and Max Coltheart. 2002.
\newblock \href {https://doi.org/10.1080/02724980244000099} {{358,534 nonwords:
  The ARC nonword database}}.
\newblock \emph{The Quarterly Journal of Experimental Psychology Section A},
  55(4):1339--1362.

\bibitem[{Russell(1980)}]{russell1980circumplex}
James~A. Russell. 1980.
\newblock \href {https://doi.org/10.1037/h0077714} {{A Circumplex Model of
  Affect}}.
\newblock \emph{Journal of Personality and Social Psychology},
  39(6):1161–1178.

\bibitem[{Scherer(2005)}]{scherer2005emotions}
Klaus~R Scherer. 2005.
\newblock \href {https://doi.org/10.1177/0539018405058216} {{What are emotions?
  And how can they be measured?}}
\newblock \emph{Social Science Information}, 44(4):695--729.

\bibitem[{Schuff et~al.(2017)Schuff, Barnes, Mohme, Pad{\'o}, and
  Klinger}]{schuff2017}
Hendrik Schuff, Jeremy Barnes, Julian Mohme, Sebastian Pad{\'o}, and Roman
  Klinger. 2017.
\newblock \href {https://doi.org/10.18653/v1/W17-5203} {{Annotation, Modelling
  and Analysis of Fine-Grained Emotions on a Stance and Sentiment Detection
  Corpus}}.
\newblock In \emph{Proceedings of the 8th Workshop on Computational Approaches
  to Subjectivity, Sentiment and Social Media Analysis}, pages 13--23,
  Copenhagen, Denmark. Association for Computational Linguistics.

\bibitem[{Strapparava and Mihalcea(2007)}]{Strapparava2007}
Carlo Strapparava and Rada Mihalcea. 2007.
\newblock \href {https://aclanthology.org/S07-1013} {{S}em{E}val-2007 task 14:
  Affective text}.
\newblock In \emph{Proceedings of the Fourth International Workshop on Semantic
  Evaluations ({S}em{E}val-2007)}, pages 70--74, Prague, Czech Republic.
  Association for Computational Linguistics.

\bibitem[{Synnaeve(2015)}]{phn2vec}
Gabriel Synnaeve. 2015.
\newblock \href {https://github.com/syhw/speech_embeddings} {{Speech
  Embeddings}}.
\newblock Github Repository at \url{https://github.com/syhw/speech_embeddings}.

\bibitem[{Traxler and Gernsbacher(2006)}]{Traxler2006}
Matthew~J. Traxler and Morton~A. Gernsbacher, editors. 2006.
\newblock \href {https://doi.org/10.1016/B978-0-12-369374-7.X5000-7}
  {\emph{Handbook of Psycholinguistics}}.
\newblock Elsevier.

\bibitem[{Warriner et~al.(2013)Warriner, Kuperman, and
  Brysbaert}]{Warriner2013}
Amy~Beth Warriner, Victor Kuperman, and Marc Brysbaert. 2013.
\newblock \href {https://doi.org/10.3758/s13428-012-0314-x} {Norms of valence,
  arousal, and dominance for 13,915 {English} lemmas}.
\newblock \emph{Behavior Research Methods}, 45(4):1191--1207.

\end{thebibliography}

\clearpage

\onecolumn

\appendix

\section*{Appendix}

\section{Best and Worst Predictions of Models on Nonwords}
\begin{table*}[h!]
  \centering\scalefont{0.9}
  \setlength{\tabcolsep}{15pt}
  \subfloat[][Trained on nonsense words, phoneme 1-gram model]{%
  \begin{tabular}{cllllll}
    \toprule
    &joy&sadness&anger&disgust&fear&surprise\\
    \cmidrule(r){2-2}\cmidrule(lr){3-3}\cmidrule(rl){4-4}\cmidrule(lr){5-5}\cmidrule(rl){6-6}\cmidrule(l){7-7}
    \multirow{10}{*}{\STAB{\rotatebox[origin=c]{90}{Best Predictions}}}
    &bange	&gnirl	&zunch	&plert	&phlump	&scrare	\\
    &groose	&drusp	&sout	&twauve	&cruck	&twale	\\
    &cisp	&shuilt	&swetch	&framn	&cliege	&gnewn	\\
    &gnirl	&scrare	&wholk	&sout	&purf	&psoathe\\
    &broin	&throoch&chuile	&gnirl	&snoob	&phreum	\\
    &chuile	&prote	&cisp	&throoch&scrol	&theight\\
    &swetch	&phrouth&framn	&theph	&chuwk	&grulch	\\
    &shuilt	&zunch	&preak	&purf	&grulch&cliege	\\
    &kass	&theight&yirp	&cisp	&twale	&thwick	\\
    &throoch&flalf	&dwull	&zorce	&ghuge	&plert	\\
    \cmidrule(r){1-2}\cmidrule(lr){3-3}\cmidrule(rl){4-4}\cmidrule(lr){5-5}\cmidrule(rl){6-6}\cmidrule(l){7-7}
    \multirow{10}{*}{\STAB{\rotatebox[origin=c]{90}{Worst Predictions}}}
    &purf &hupe	&snusp	&ghuge	&bange	&blidge \\
    &snoob	&snoob	&broin	&grulch	&phreum	&zel	\\
    &cruck	&phype	&shrote	&slanc	&gnirl	&cheff	\\
    &plert	&broin	&blidge	&shrote	&snusp	&dwull	\\
    &snusp	&dwear	&slanc	&groose	&psoathe&purf	\\
    &skief	&wholk	&phrouth&thwick	&phrouth&ghuge	\\
    &yirp	&skief	&plert	&hupe	&broin	&throoch\\
    &slanc	&slanc	&scrol	&cruck	&pseach	&snoob	\\
    &choff	&sout	&skief	&fonk	&slanc	&cisp	\\
    &yourse	&preak	&shuilt	&theight&chuile	&pseach	\\
    \bottomrule
  \end{tabular}
}

\subfloat[][Trained on real words, character 2-gram model]{%
  \begin{tabular}{cllllll}
    \toprule
    &joy&sadness&anger&disgust&fear&surprise\\
    \cmidrule(r){2-2}\cmidrule(lr){3-3}\cmidrule(rl){4-4}\cmidrule(lr){5-5}\cmidrule(rl){6-6}\cmidrule(l){7-7}
    \multirow{10}{*}{\STAB{\rotatebox[origin=c]{90}{Best Predictions}}}
    &blidge	&slanc	&blour	&phype	&tource	&sloarse\\
    &wholk	&theph	&drusp	&twauve	&twarp	&preak\\
    &yirp	&zel	&plert	&twale	&grulch	&phrouth\\
    &cheff	&twauve	&ghuge	&phreum	&yirp	&gnewn\\
    &hupe	&bange	&zant	&fonk	&sout	&choff\\
    &shrote	&valf	&wholk	&yourse	&swetch	&phreum\\
    &dwull	&cliege	&rhulch	&zerge	&cliege	&glelve\\
    &gnewn	&grulch	&cruck	&scrare	&scrol	&cruck\\
    &framn	&phrouth&snoob	&gnewn	&sloarse&grulch\\
    &yealt	&gnirl	&gnirl	&scrush	&dwull	&psoathe\\
    \cmidrule(r){1-2}\cmidrule(lr){3-3}\cmidrule(rl){4-4}\cmidrule(lr){5-5}\cmidrule(rl){6-6}\cmidrule(l){7-7}
    \multirow{10}{*}{\STAB{\rotatebox[origin=c]{90}{Worst Predictions}}}
    &snoob	&ghuge	&blidge	&valf	&phrouth&zel\\
    &theph	&phlump	&broin	&shrote	&prote	&throoch\\
    &thwick	&chuick	&valf	&scrol	&snusp	&twale\\
    &chymn	&prote	&chuile	&phrouth&chuile	&chymn\\
    &snusp	&chuile	&swetch	&skief	&psoathe&scrare\\	
    &preak	&zunch	&snusp	&dwull	&cheff	&purf\\
    &swetch	&purf	&phrouth&zunch	&shuilt	&kass\\
    &twale	&yealt	&zorce	&prote	&chymn	&twauve\\
    &yourse	&swetch	&sout	&chymn	&bange	&bange\\
    &cisp	&choff	&tource	&ghuge	&broin	&snusp\\
    \bottomrule
  \end{tabular}
}
\caption{The top 10 best and worst predictions for nonsense words by
  the best model trained on nonsense words and the best model
  trained on real words.}
\label{non_top_10}
\end{table*}

\clearpage

\section{Excerpt from our lexicon of nonsense words with emotion
  intensity annotations}
\label{sec:corpusexcerpt}

\begingroup
\small\sffamily
\begin{tabular}{lllllllllll}
  \toprule
IDs & Word & ARPA Pron & Real & Joy & Sadness & Anger & Disgust & Fear & Surprise \\
\cmidrule(r){1-1}\cmidrule(rl){2-2}\cmidrule(rl){3-3}\cmidrule(rl){4-4}\cmidrule(rl){5-5}\cmidrule(rl){6-6}\cmidrule(rl){7-7}\cmidrule(rl){8-8}\cmidrule(rl){9-9}\cmidrule(rl){10-10}\cmidrule(l){11-11}
0 & afraid & ah f r ey d & 1 & 0.3125 & 0.8333 & 0.3333 & 0.1875 & 0.6875 & 0.3333 \\
1 & alse & ae l s & 0 & 0.6875 & 0.4375 & 0.5625 & 0.4792 & 0.4375 & 0.5625 \\
2 & apache & ah p ae ch iy & 1 & 0.2917 & 0.6458 & 0.7708 & 0.4792 & 0.5 & 0.5833 \\
3 & aphid & ae f ih d & 1 & 0.3333 & 0.625 & 0.4792 & 0.5625 & 0.6042 & 0.3125 \\
4 & bale & b ey l & 1 & 0.5 & 0.5208 & 0.4167 & 0.3542 & 0.4583 & 0.0833 \\
5 & bange & b ae n jh & 0 & 0.375 & 0.4375 & 0.6458 & 0.6042 & 0.7917 & 0.75 \\
6 & battle & b ae t ah l & 1 & 0.1667 & 1.0 & 0.9583 & 0.7083 & 0.7292 & 0.5417 \\
7 & bias & b ay ah s & 1 & 0.2292 & 0.5625 & 0.5625 & 0.4167 & 0.5417 & 0.4375 \\
8 & bizarre & b ah z aa r & 1 & 0.4583 & 0.625 & 0.6042 & 0.5417 & 0.4792 & 0.5833 \\
9 & bleve & b l iy v & 0 & 0.4792 & 0.4167 & 0.3125 & 0.375 & 0.4167 & 0.5417 \\
10 & blidge & b l ih jh & 0 & 0.6042 & 0.4375 & 0.7083 & 0.4583 & 0.6042 & 0.7292 \\
11 & blister & b l ih s t er & 1 & 0.4375 & 0.625 & 0.4375 & 0.625 & 0.7083 & 0.4583 \\
12 & blour & b l aw r & 0 & 0.4583 & 0.5833 & 0.4375 & 0.4167 & 0.3125 & 0.6042 \\
13 & blurnt & b l er n t & 0 & 0.5 & 0.4375 & 0.3542 & 0.3958 & 0.3958 & 0.5 \\
14 & blusp & b l ah s p & 0 & 0.5417 & 0.5417 & 0.6458 & 0.5208 & 0.4583 & 0.4792 \\
15 & boarse & b ow r s & 0 & 0.2708 & 0.6875 & 0.7708 & 0.8542 & 0.8542 & 0.5417 \\
16 & boil & b oy l & 1 & 0.2708 & 0.75 & 0.75 & 0.3958 & 0.3958 & 0.3333 \\
17 & bowels & b aw ah l z & 1 & 0.0833 & 0.5208 & 0.4792 & 0.8333 & 0.5 & 0.4583 \\
18 & break & b r ey k & 1 & 0.6875 & 0.7917 & 0.6458 & 0.3125 & 0.2917 & 0.4792 \\
19 & broil & b r oy l & 1 & 0.25 & 0.7083 & 0.875 & 0.75 & 0.7917 & 0.3333 \\
20 & broin & b r oy n & 0 & 0.375 & 0.6458 & 0.8125 & 0.5833 & 0.6875 & 0.5208 \\
\ldots \\
319 & whalk & w ae l k & 0 & 0.6458 & 0.3333 & 0.2708 & 0.3125 & 0.5417 & 0.5625 \\
320 & wheuth & w uw th & 0 & 0.6875 & 0.4375 & 0.5 & 0.5417 & 0.5208 & 0.625 \\
321 & whoal & w ow l & 0 & 0.6458 & 0.4375 & 0.3333 & 0.375 & 0.3542 & 0.7292 \\
322 & wholk & w aa l k & 0 & 0.3958 & 0.625 & 0.5 & 0.5417 & 0.5208 & 0.5833 \\
323 & wrause & r ao s & 0 & 0.4792 & 0.4375 & 0.6875 & 0.625 & 0.5833 & 0.5208 \\
324 & wrelt & r eh l t & 0 & 0.5833 & 0.5208 & 0.5 & 0.4375 & 0.4375 & 0.3125 \\
325 & wrilge & r ih l jh & 0 & 0.625 & 0.5208 & 0.4792 & 0.5833 & 0.625 & 0.5 \\
326 & wrorgue & r ao r g & 0 & 0.3125 & 0.5417 & 0.7083 & 0.625 & 0.8542 & 0.4375 \\
327 & wruse & r uw s & 0 & 0.4792 & 0.6042 & 0.5417 & 0.5417 & 0.6042 & 0.625 \\
328 & yage & y ey jh & 0 & 0.3542 & 0.625 & 0.625 & 0.5833 & 0.6667 & 0.4583 \\
329 & yealt & y iy l t & 0 & 0.3542 & 0.5208 & 0.4583 & 0.4167 & 0.6458 & 0.4375 \\
330 & yirp & y er p & 0 & 0.4375 & 0.5625 & 0.4167 & 0.5417 & 0.4167 & 0.5417 \\
331 & yourse & y uw r s & 0 & 0.6458 & 0.3542 & 0.25 & 0.3333 & 0.5208 & 0.5208 \\
332 & yurch & y er ch & 0 & 0.5625 & 0.5 & 0.4792 & 0.5208 & 0.4583 & 0.5625 \\
333 & zant & z ae n t & 0 & 0.5417 & 0.3542 & 0.4375 & 0.4792 & 0.4792 & 0.5 \\
334 & zany & z ey n iy & 1 & 0.7708 & 0.0625 & 0.2708 & 0.3542 & 0.125 & 0.5417 \\
335 & zel & z eh l & 0 & 0.6667 & 0.375 & 0.5417 & 0.2083 & 0.3958 & 0.75 \\
336 & zerge & z er jh & 0 & 0.6667 & 0.3333 & 0.4375 & 0.4167 & 0.4375 & 0.5625 \\
337 & zorce & z ao r s & 0 & 0.4583 & 0.5833 & 0.7917 & 0.6667 & 0.625 & 0.625 \\
338 & zourse & z ow r s & 0 & 0.5625 & 0.3958 & 0.5833 & 0.5208 & 0.375 & 0.6458 \\
339 & zunch & z ah n ch & 0 & 0.4583 & 0.6667 & 0.625 & 0.7292 & 0.7083 & 0.4375 \\
\end{tabular}
\endgroup

\clearpage

\section{Complete Nonsense Word Emotion Intensity Lexicon}
\label{sec:base}
Copy-paste the following character sequence into a plain text file
with the name \verb|data.txt| and execute
\verb#base64 -d < data.txt | bzcat > nonsense-words-emotion-intensities.csv #
\vspace{5mm}
\begingroup
\tiny
\begin{verbatim}
QlpoOTFBWSZTWdlU38YAIfpfgEAQQAF/6CUwWIA///+wYB3wPY1UFBfefHd1y7Z2WwdmpRkdpzhgF2zo63GoO7KoaN
ngAAAAAAgAAfRuwRvttc4gPumOpUnGGmhB6iBMgmhU09TKNmpMnkGmCRRBNEmnqeUMgAABJlSKSfooADIaAAABJ6pU
JMSgAMgAAABo0KTRoGgAyB6jQNNASahETRJkamTRoAAAOZgzP2ec8Bz1cRKPswRYFaYoY0bIsJSKNlLgbQSQD9NZOx
XQgJaC2o7QVhjaqjBkScQOGKwSyq2o2RRJFNQohe5ocTsDEZEhiRbYYZabmu4CJEkii8bIsITyAiNZCC4zEiFIwERG
gjigibYSPG6YTcbgYSEhZJCJQljM7qoYlYEKxLO7EqXaqmV3LtbL3BN17zZPNbN3/L5Ir/X+75pf//StJTrV50QeC7
/nR9BmjY1sZkPojGpFA7yDpvX+sacZ20nNtp4NNXOuloNNLTvpWs0zqXjekbOKYpFTvauaZMGp3NzBsZtvKVuGOdbe
vnq9z6rtc+44sO3/EM4+UNRh6pVw/8qwwA44gbUc3NCooX7OLZrzKe0YMbzZGxjB4V8S3Hp8xx3eeeYiFPp49b6PXo
T58/pWjruopCGq/KzxYVk9uwR3rmUHMDa2VY0qVKl97QLmj1t2mb0j21DyfUj4MPjnnhb6tr0kfSTYvjoFIAY+CF7h
cvOnfHNBbPOObE17fBBPCPbbbRWpnVpFuPfTJDfddHsdIJ7zEwcPkryX7Xm0BatKxeuJF4gcKOIevsfNEHere3u01Q
hTo+SQmdT4p6W+DqubA/CiJTIKTp1IgivW99ab6512gXvvzN7dqsjZ3xLvsetwGBaRix4p2aEdWqA+o7+CPwfj5JQQ
QKxp0wfP2/ALRu/NcBrN0ib10TxJO77d84RtOjyKNGlBJBBAy7zx4nSOlHpGv4DCICM9UQfkeJpxBw/AI1KKYN+ZUH
SHqWfHSANlVXeipeGyJNcfgzZYG9Z73PNK6dO9HgoEnqvEfXnauEeLCmxg4kDBrBkW1pqZrQVOXW2pk6kSYjRVZjlL
dU1XjTk8PPWPs+4hPqFSEhJJ+OMbbbX1fTV9+5zlXA/HbdONhcC4d11V0BznKn7BBcZBcr0AkguViUiqTYKEgSJiRl
KGijbBkoizNGo0GjYBsnnvPnx48e3lo2LRABmYlgD/ilKft/dL8frokkkkkkkkkkkjXRu4eIMm9buQu7LgwztKhTRV
UMSqWRiymazFRLla0IqzScxx71dyF3dw1AEmZlyF3dkyQfBCB0wHV7e3t6r1ExizN6tN0GCiGksloMaSxFWSJSIoii
xFGta1cmyAhmZchd3yEAJmZrppnOdEkkkkkkkkkklKYkEkEkjz1whESjKQRTT5q+SB4hDmiLIEFJc3zzmSGZlwNhBM
zLkLu7kLu67mmotNEUFRigsEezFPKJXTVIsCKUoxbKJN3XbGZbhVxMYNGqIxGKjbu4rJyk3SJd3SJpLI4aImBEQR49
982Oc52AGdkhrWrkLu++bkAamMiCKoixJNCnavXj1876YkxoiQSZeVw0oZKAsIJPft2TQMADMoFApSFJgUYKEowZNM
mikMhsSJlEZHt59vbz8fT3+vxlJJJJJJJJJJJMGZqEgliWYsSSQQsVEFVHqQzLs7d4BGoGtauQu7CSbgoBkqQkrchI
kC5EtiqoqrRSYo1GZsaiN3r159Xnz5zISBmZmZmZ3DTFBRioip2Kpipmd3RDuuSaBEEySDBgyJC51Gw7ulIApFUUFi
xTe973t2b3ve5IQ61rWp74wZhZgxCozI0RgTTDIElCWIlopZKNJGkxEliSgwIJBYkEkEsCxBIBBmZmyze+MpJJJJJJ
JJJJLERlmAYYAYVBZgSAwJYMGpSnjyzIWRsL11ySwmQDIIaRKEzQiBMwiaZEwpppmaHr169atb2vPnqrzA72wbmnLk
FFxsuByI7qmGiIlgzBoiHd3d7REVzS2c5ykkkkkkkl48ePHjx48izkF9KBWJoEalzK7UOkVyqNWqFq0U6EclXprKqs
gEnWZmF3dgQm4B2USCgCwFiyCwFSa1rV3dkhDrMzLu+/dEiAxptvW2tObWi2oqJLKDDEpSKilQjGZkiEwRhiRjKJlG
evXr148eL5kZISwWRgHnqqnKKlEoiyEiZcM1KoLOVcuYYUqXZBWgtEELmoVUGrNK2mpnKWpqqSZs0FpJGlnJVlRFUY
qFiVYdRIlmGFs4ZdDpGmaYQZGpBiaWBElrKDJLWmmIqGhiXOGrILWZpUUctRNmRWnTKCIgxM2GMd3GUnd1BDs5xZ3J
dnImXVsK2MZSSSSSSSSSSSWQwYPTrqrz6BFEHCuRfIc1WnUENirjuSWFkaTAkMJKaZGJAc4YyRRRda1VVkhJvk1rWr
u7JAOpC0IqyEWMUaMk8vLeKEiaJCTRmYymwoaQjSyxDCg1LIaMbERt47x5tTREO75GDMzeVfOFlJJJJJJJJJJJWgFp
YZoEhSjKZ9PJ9dtXcRiRQisGCoLDvqyqrrRCZmXd3VXqx5jjtlSahSRSHLOQQVkjKNEio4ETIiJUaQsgzLpxCBILss
gLMi7TtCmC5O7RCZy7MsmC5cVOci0laoF3ffdXYG5zlXAD01r6q1itS9evTXzNIooiElkILTBgikFIxslGImjMwRpp
JjYTIzCgpNMxIGGSJQGAE0mSjx3j39vj2+fn4+vx8JJJJJJJJJJK5MsSCCCQQCCQWJBgiqLDrWu8hLu6qsk0qqDBiA
IvW1dXa1G2yZGjQZtCJgJNAUVk0UWTUxFIsVRFUXWtVVaJN5mYXd9VjU7jckRy5znNgLlYiNzFWjHKqsC7vfAxRgoo
qKxUBJRJiEohMvp2uYlQhkh3boUmNJGTBREUEQgY0SJMUAmSc4DJFFANhACCSQQQSxBIIpSlM0vnOcJJJJJJJJfHx8
fHx8fD58d9Zr3VERsSNKYbz57eEiA2ZFhmQIjEUEVixslSEaNUkKiRFLusJvrWtVVcQOyBxEJKQFgDEsBSlIiIoBEQ
7u7u7vObZws5SSSSSSSSSSSVRMVFyTRi9IehFQWNYdnr+tKyz1Gxw8Ka4K2FtkioQE1oihHBy4hpukNpFomHi5WRNH
Dk0c1MPTQkCeBuQ6DogpNHOPOc27bNZa0h18Tg0aEyG0RAXGYGUbuaA+XjCI4XYkG2UiiF5nLLIERGp0NUWwOcKQ7F
DAN1R1CpFwUHgRLILMYUqdFGx4QkCQxxwTUUGCxSgbaCyFHq0KOMIOpw2LkuOt3SIy25WLMRA40HSWQxqc2kqkSxgH
h1kzTsCCBNoVbwogkAluMMMqKPDKnGGExVZDu2QguTeXZoVYQTILgsd402K0xrjNctYoOIgI0hVGYoWa0HEQZpg5WN
0Uw8NLIpCAMXLbqb2bHHJApbu21G7FJIShII0KzWUoK4I6nvHIt1i8R3aggntBIaRBd4EEbayQWKbENqEk0QrG1alU
TIqIVSZZI6lELyqzSSzDE3skUMnFA0qXRrG7Xk3SqVVzQFpr3UKcOmvEt0icQQvKnrVhZiSChupwg7hQIreMMU43bS
JkShKQRXNFTBocuAk1ThrhSjClhgcvNGu6iSiGNIZogYYQIIQaBlaFVrrIooKQEKaJLohxRcL0RFpvWnAdeWpFmEa6
1pWs0MMmA1Fmya0LlaxxsEiNDTu6cSdTpZEAJO16RBAW0jUKDUWVMJpNOApBQ1hWSOJESSQMV2BJERCKy7c3GktQ0z
NVYIqbtZ1CHnG0eHzGTyDuaTGbEIpSgYX1LY0rtFEBFYQZlPXle6goHoWwIbpQRxnKkmztsJqgxBOxxIOV5DSC3XYE
bkhTbTbKRUYjcCqyViuMx6WFFKtFqLa7bvGuZAdlpj4pprw4UWXG8JLm94gXNDUBRWwN1MPjDJGxyLNQvYxEghV3Q5
3sIOxgwJoIcLrDeVimxlmEkkklnWZDo2aHgfEN2WYcfEiIaykci0yBGA6Ai9cVSqFIIkSHEnCNRjRO8GrUCFuuHYVE
ncUo1OkTQi0VdArBu7A+uWijSQ0+hBjdFfRwkIzRQzrEQNI3TLkMKDHGpiHKiEIuGGmWiloSFCiLEzx59fz/R9kfsP
y+rpn79n36hfsuwfmXXsClNizUTfJjvl3xzvKIc7954/d4308bD+YNaQ5zXPK/I95pffTs96tnlLTaQV72Dr9KLel3
NYEbP/iyxZYxOmudUtnFiSFbIrM6i6vYGtI0mRXSRCkK85pSJu4N1WLqLEvYg20zF8a7a0WXsdVEuopqbTYRoIwhML
8Y2GrjS4478+LU540ONunzPINLReavDVt1R6gmhsAcIO7xzKzxdG2lqzaytY5RdaYU0r+PLv5PPw7VB2vpDkb8bvt9
d/rx8/fXPPKSSSS+hzrqNuN9999999990kln3pzx37yIj466cVhPYRd6ddddddddddJJJJJJJJLON+eJ8D6fj169Xu
1YqXE+ZtL+HFfHjx48ePHjx4SSWunvtzPP4MCvYitfiZES8nnnnnnnnnnlJJJJJJJJJJJJLA3IznOc1Gc/7U/T7bbR
9e/v5n7IBI8Vfv379+/fv344SS0611222222+LAfGM8ccccbbbL7+e/O+/kOA/XXbt2r9HXy7gjXR+uuuut99990kk
kkkkkkkkkktfvw+oILdjtttttttt4XGmo5n6hyYmkeKN4j6IvLuCLJ7zz1IsKO5/tppSaKmPs5/tP3dVUB/fv379+/
fv37SXW3vXfcc/H1DU5555555555SSSSSSSSSSSSSSSSSSSSSSSSSSXGnBbv67R5LcGvb26uKKTXxaHrNeuuuuuuuu
uySSSSSSSSSSS539a66668Deo457du3bt22222Sx212Ouuuutb22Pwd99999tvz+rfwf8/s/+o/dkKyAAFPxwABX0f
jo8Cp7Z0oUIFVEAiTRZzo3/tMAXIDMSzAasv3P6KwHY7bfvS38X/fI1rgWBmH/mfAwtfyOA2SABu9zpkXfTUuV5Xj+
Hq1ednzagtWzUIO9Jt+T6mY8W5YADrkM5AYsBbF/ffXit2xFaTmZ5CTcAuw3B5OEDB9eHvj+iSnTeyDXX5zfYDy86w
xlHOB4q7RFvK7ZPCIqqkAQA8tY+DaowdZm4Y4pJVnRAYYgMAD5076XL/FYZF+J8084FBe1r9RXatdL5VLseopOtLw4
75nD0F0VtKFnE4mHOYoVjFFWuYFxhKJNoyaQ75KrR3ES32fPy4/5HPDm3le6fXrHiflZpQfLACxZhQScc/U96ZxIVH
xvbHT6BmZ/Ntrvr5yK59Vn4n3t2FW8+M35wh6QRFJBYwD5jTx5+9+Tfy1rfNcJxvAFIcodkSIUhGUyytZK9evdWr3n
WVWYPHCaO1QtmrokrXRZ4YfL2W/Gn5k8uyEeUcXLecn4lXedwO86pdBCSAIBCBQQXJLFiRJpwiXl6jvrrrt7RBVrHB
qlGMEppQzWmsGTjRd55rzXFUshGBZC2goo8jCLLMckQighCU1Hrx3dFMrqsSsCXyxxOfIkcyonQSPv7Wa8GurSANmu
Qvf00rQNEOIEoKAVkw9qVhmpEd+1/PUV6D95b077c9+dsRbayideO2Bp43z5AYDECdCa743IjGuu+JyFmX9adhhfKF
tc/88+nfj028fC6909WQnw9cwDABgiiIrFYosWMZCa6zn23HM69+ZqtVo5PoQ4ZF47UkigsiyCwgGTrdPO09+X31d7
vy9fl8+5nzsMNnq2nmaHEqaqhnHVBa09qcxFmmuL3HEw7nbTRMcA4gXRuS5dw6gTUWsL2GLU5LEksQzDtB500m9LBp
vC4tTZA3Dib0kDLEglgSQCxIDEkszRHpVfJ0fwDaF6uLtSuRiL/FjN40ACH6LGdrY0rlVpV50qK2mZea1FiIADcixH
PPxPi33rt61gW9T8awc0fVx5OPO9W1ppI3wO1LFj6qR9H+RpXqo26302HRrry/jIwQavW1e81iI3+e1J3LUtx8sAWr
z+Z9PG3/fXaCKSIiI1b3oPR+pca+TVRhvPX7PvfvohGDTgjSzqRJInnuEe7k2JvnXKLJZNX0lc2jo7rFrsyy13XJtB
QHIJZyASWBBBNk5dwod+3vvz6MY71l49RTvnQ5BLb+YkeNgtsw2wzTQw+z43LHbdpvLq2I1udmtiJcUvfGwf9qbxpr
xxi1DxL2NH41O5d2Yg7xxBZedtnHPkeh6ee/YePL2e3mLFgoRBhFkAsCxLAlmLBgOf0827SOV4gQgwpbvc0MRE3t3N
64rVYnr2EeUk8EOeHfG9knG6JKHDrVXwkDMd6pdU5oDbJzFVERUZEAbHvjRydO+cHXS9lR5uyESdcVgYrLvYzTDXNX
iHt4hDwQvttfdR4txDcIe3kpGKrdQp/qlgTT7nyxJPE1au5lFqmtZpMcW1o9bRudVft9fjCpm2k98x4xMXTv29frLY
w5qbH+/QmtWHL+mZm7aeRaMJ8dXvmd8qvjvp4nKgpnFBbNYOUUotzGF3KhaZmIrVLQxpozCrBUYFiBxIJBEkIuRn6+
PqPn78/GKYmmhzkgE1pTmqwi7grvjwPQHwgSOBlIMJpANLEs4EmUQ20SmyES2UGSRcqDBuHLnPazYy5k6xksxDd2cY
IgFnWX/6LuSKcKEhsqm/jA==
\end{verbatim}
\endgroup

\end{document}